# A computational scheme for reasoning in dynamic probabilistic networks

Uffe Kjærulff
Department of Mathematics and Computer Science
Institute of Electronic Systems
Aalborg University
DK-9220 Aalborg, Denmark
uk@iesd.auc.dk

## Abstract

A computational scheme for reasoning about dynamic systems using (causal) probabilistic networks is presented. The scheme is based on the framework of Lauritzen and Spiegelhalter (1988), and may be viewed as a generalization of the inference methods of classical time-series analysis in the sense that it allows description of non-linear, multivariate dynamic systems with complex conditional independence structures. Further, the scheme provides a method for efficient backward smoothing and possibilities for efficient, approximate forecasting methods. The scheme has been implemented on top of the HUGIN shell.

## 1 INTRODUCTION

The application of probabilistic graphical models (belief nets, influence diagrams, etc.) for modeling domains with inherent uncertainties has become widespread. A common trait of the domains, where such applications turn out most successfully, is their static nature. That is, each observable quantity is observed once and for all, and confidence in the observations remaining true is not questioned. However, domains involving repeated observations of a collection of random quantities arise in many fields of science (e.g. medical, economic, biological). For such domains a static model is not very useful: the estimation of probability distributions of domain variables based on appropriate prior knowledge and observation of other domain variables is reliable only for a limited period of time, and further, upon arrival of new observations, both these and the old observations must be taken into account in the reasoning process. Thus, to cope with such dynamic systems using probabilistic networks we need to interconnect multiple instances of static networks. Obviously, as time evolves, new 'slices' must be added to the model and old ones cut off. This introduces the notion dynamic probabilistic networks (DPNs).

In general, a dynamic model may be defined as a sequence of submodels each representing the state of a dynamic system at a particular point or interval in time; henceforth, such a time instance will be referred to as a *time slice*. Hence, a DPN consists of a series of, most often structurally identical, subnetworks interconnected by temporal relations. To make estimates of variables of a dynamic system in a way that makes full use of the information about past observations of the system, requires a compact representation of this information. The creation of this representation is part of the process of reducing the dynamic model. This reduction process includes elimination of parts of the model representing past time slices, and should have no effect on future estimates, that is, the information conveyed by the eliminated part of the model should be completely represented in the remaining part. The complementary process of expanding the model must be carried out whenever new time slices have to be included in the model.

In classical time-series analysis (see e.g. Box and Jenkins (1976) or West and Harrison (1989)) the emphasis is on model assessment, i.e. estimation of model parameters given a time series of observations of some stochastic process. The model thereby selected is then used for making predictions about future behaviour of the time series. Although the classical time-series analysis techniques have been quite successful, their ability to cope with such important issues as complex independence structures and non-linear relationships of have appeared to be rather modest. By formulating the analysis in terms of DPNs both of these limitations vanish. Attempts to integrate methods of classical time-series analysis with network representation and inference techniques have been presented by Dagum, Galper and Horvitz (1992). This paper, however, does not address the issue of model assessment, but merely problems related to making inferences (including prediction and backward smoothing, in classical time-series analysis terms). That is, the dynamic model is assumed to be given.

Among research activities applying DPNs, as defined above, are a model for glucose prediction and insulin dose adjustment by Andreassen, Hovorka, Benn, Ole-sen and Carson (1991), an approach to building planning and control systems by Dean, Basye and Lejter (1990), a model for making judgements concerning persistence of propositions by Dean and Kanazawa (1989), and a model for sensor validation by Nicholson and Brady (1992). However, none of these activities have dealt with the issues of reasoning in DPNs.

In Section 2 we briefly review some relevant graph theoretic concepts as well as some fundamental characteristics of conventional (static) probabilistic networks and some of the DPNs introduced. The processes of reducing and expanding DPNs are described in detail in Section 3 as well as the processes of backward smoothing and forecasting. Section 4 briefly summarizes the presented scheme and provides a list of some of the yet unresolved issues.

## 2 TERMINOLOGY

Commonly used graphtheoretic terms like 'directed graph', 'undirected graph', 'triangulated graph', 'parent', 'children', 'cliques', 'paths', 'cycles', etc. shall be used without formal definitions; see e.g. Lauritzen and Spiegelhalter (1988) for details on relevant the terminology. We shall use the following abbreviations: the set of parents, children, ancestors, and neighbours of a vertex $\alpha$ are denoted by, respectively, $\text{pa}(\alpha)$, $\text{ch}(\alpha)$, $\text{an}(\alpha)$, and $\text{adj}(\alpha)$. In the sequel the symbol $\otimes$ denotes the binary operator producing the set of all unordered pairs of distinct elements of its arguments. In the following two paragraphs we review some less common graphtheoretic notation.

For a directed graph $\mathcal{G} = (V, E)$, $\mathcal{G}^m$ denotes its *moral graph* obtained by adding edges between pairs of vertices with common children and dropping the directions of the edges. A *decomposition* of an undirected graph $\mathcal{G} = (V, E)$ is a triple $(A, B, C)$ of non-empty and disjoint subsets of $V$ such that $V = A \cup B \cup C$, $C$ separates $A$ from $B$, and $C$ is a complete subset of $V$ (i.e. each pair of vertices in $C$ are neighbours). A decomposition $(A, B, C)$ *decomposes* $\mathcal{G}$ into subgraphs $\mathcal{G}_{A \cup C}$ and $\mathcal{G}_{B \cup C}$ (i.e. subgraphs induced by $A \cup C$ and $B \cup C$, respectively). $\mathcal{G}$ is *decomposable* (triangulated) if and only if $(A, B, C)$ decomposes $\mathcal{G}$ and both $\mathcal{G}_{A \cup C}$ and $\mathcal{G}_{B \cup C}$ are decomposable.

When a vertex $\alpha \in V$ and the edges incident to $\alpha$ are removed from $\mathcal{G} = (V, E)$, $\alpha$ is said to be *deleted*, but when $\text{adj}(\alpha)$ are made a complete subset by adding the necessary edges (if any) to the graph before $\alpha$ and the edges incident to $\alpha$ are removed, then $\alpha$ is said to be *eliminated*. Note that connectivity of a graph is invariant under elimination, but not necessarily under deletion. The set, say $T$, of edges added by eliminating all vertices in $V$ in any order is called a *triangulation* of $\mathcal{G}$ as $(V, E \cup T)$ is triangulated. The edges of $T$ are called *fill edges* or *fill-ins*. An *elimination order* is a bijection $\# : V \leftrightarrow \{1, \ldots, |V|\}$. $\mathcal{G}_\#$ is an *ordered graph*. The triangulation $T(\mathcal{G}_\#)$ is the set of edges produced by eliminating the vertices of $\mathcal{G}$ in order $\#$. An elimination order $\#$ is *perfect* if $T(\mathcal{G}_\#) = \emptyset$.

A probabilistic network, as used in this paper, is built on a directed, acyclic graph (DAG) $\mathcal{G} = (V, E)$, where each vertex $\alpha \in V$ corresponds to a discrete random variable $X_\alpha$ with finite state space $\mathcal{X}_\alpha$. For $A \subseteq V$, $X_A$ denotes the vector of variables indexed by $A$. Similarly, $x_A$ denotes an element of the joint state space $\mathcal{X}_A = \times_{\alpha \in A} \mathcal{X}_\alpha$. Each random variable $X_\alpha$ of a probabilistic network is described in terms of a conditional probability distribution $p(x_\alpha \mid x_{\text{pa}(\alpha)})$ over $\mathcal{X}_\alpha$, where $p(x_\alpha \mid x_{\text{pa}(\alpha)})$ reduces to an unconditional distribution if $\text{pa}(\alpha) = \emptyset$. In $\mathcal{G}$, the conditioning variables of $X_\alpha$ are represented by $\text{pa}(\alpha)$. The joint probability, $p = p_V$, over $\mathcal{X}_V$ is the product of all conditional and unconditional probabilities. $\mathcal{G}$ is called the *independence graph* of $p$, since for each non-adjacent pair $\alpha, \beta \in V$, $\alpha \perp\!\!\!\perp \beta \mid \Gamma$ if and only if any path between $\alpha$ and $\beta$ in $\mathcal{A}^m$ contains at least one member of $\Gamma \subseteq V$, where $\mathcal{A}$ is the subgraph of $\mathcal{G}$ induced by $\{\alpha, \beta\} \cup \text{an}(\alpha) \cup \text{an}(\beta)$ (Lauritzen, Dawid, Larsen and Leimer 1990). Let $\mathcal{V}$ be a set of non-empty subsets of $V$. Then $p$ has *potential representation* if

$$p(x) = z^{-1}\psi(x) = z^{-1} \prod_{A \in \mathcal{V}} \psi_A(x_A),$$

where $\psi_A$ are called *potentials* and $z$ is called the *normalization constant*. In particular, the product of all $p(x_\alpha \mid x_{\text{pa}(\alpha)})$, $\alpha \in V$, is a potential representation with normalization constant 1.

By exploiting the conditional independence relations represented by $\mathcal{G}$, the joint probability space, $\mathcal{X}_V$, may be decomposed into a set of subspaces $\{\mathcal{X}_C\}_{C \in \mathcal{C}}$, where $\mathcal{C}$ is the set of cliques of $(V, E \cup T(\mathcal{G}_\#^m))$ (Spiegelhalter 1986, Lauritzen and Spiegelhalter 1988), such that computation of marginal distributions can be done in a *junction tree* $\Upsilon = (\mathcal{C}, \mathcal{E})$ (Jensen 1988, Jensen, Lauritzen and Olesen 1990) with nodes $\mathcal{C}$ and arcs $\mathcal{E} \subseteq \mathcal{C} \otimes \mathcal{C}$ representing clique intersections, where for each path $\langle C = C_1, \ldots, C_k = D \rangle$ in $\Upsilon$, $C \cap D \subset C_i \cap C_j$ for all $1 \leq i \neq j \leq k$. The existence of a potential representation is guaranteed in a junction tree, and the tree is said to be *calibrated* if $\psi_C(x_{C \cap D}) = \psi_D(x_{C \cap D})$ for all $x_{C \cap D} \in \mathcal{X}_{C \cap D}$ and all $C, D \in \mathcal{C}$, where $C \cap D \neq \emptyset$. Two junction trees $\Upsilon_1 = (\mathcal{C}_1, \mathcal{E}_1)$ and $\Upsilon_2 = (\mathcal{C}_2, \mathcal{E}_2)$ with non-empty and complete intersection $S = C_1 \cap C_2$, where $C_1 \in \mathcal{C}_1$ and $C_2 \in \mathcal{C}_2$, are said to be *jointly calibrated* if both $\Upsilon_1$ and $\Upsilon_2$ are calibrated and $\psi_{C_1}(x_S) = \psi_{C_2}(x_S)$ for all $x_S \in \mathcal{X}_S$. Calculation of marginal distributions in a junction tree is done in a two-stage process involving *collection* and *distribution* of marginal potentials between all neighbours in the tree. These two operations performed in sequence are jointly referred to as *propagation* (or *fusion and propagation*).

A DPN represents a finite (though possibly varying) number, say $n$, of time slices. Thus, the vertices $V$ of the graph $\mathcal{G} = (V, E)$ of the network consists of disjoint subsets each representing the random variables

$X(t)$ of a particular time slice $t$. That is, for some appropriately chosen $t$

$$V = V(t-n+1) \cup \cdots \cup V(t).$$

The time slices of a DPN are assumed to be chosen such that the DPN obeys the Markov property: the future is conditionally independent of the past given the present. Formally this may be written as

$$X(0),\ldots,X(t-1) \perp\!\!\!\perp X(t+1),\ldots,X(t+k) \mid X(t)$$

for all $t > 0$ and $k > 0$. Time slice 0 is called the *initial time slice*.

The set of directed edges

$$\{(\alpha,\beta) \mid \alpha \in V(t-1), \beta \in V(t)\} \subseteq E$$

is called the *temporal edges* (or *temporal relations*) of time slice $t$ and express conditional independence assumptions between slices $t-1$ and $t$. Thus, temporal edges are those between vertices of adjacent time slices (see Figure 1).

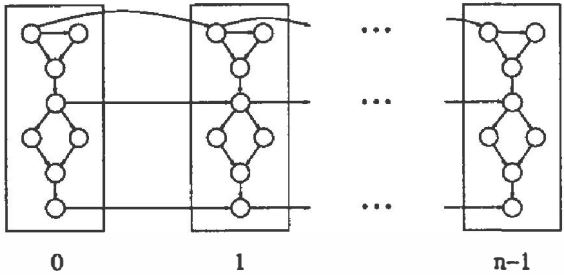

Figure 1: Sample initial DPN DAG.

At time slice $t > 0$, a DPN represents $\pi$ "past" slices and $\phi$ "future" slices (see Figure 2). Thus, the vertices

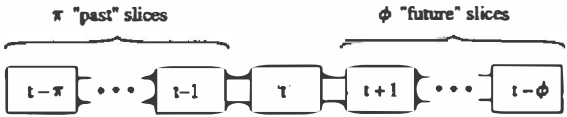

Figure 2: Time slices of a DPN.

$V$ of the corresponding graph $\mathcal{G} = (V, E)$ is given by

$$V = V(t-\pi) \cup \cdots \cup V(t+\phi), \quad \pi \geq 0, \phi \geq 0. \quad (1)$$

and the edges by

$$E = E(t-\pi) \cup E^*(t-\pi+1) \cup \cdots \cup E^*(t+\phi), \quad (2)$$

where

$$E(t) \subseteq V(t) \otimes V(t),$$
$$E^*(t) = E(t) \cup E^{\text{int}}(t),$$
$$E^{\text{int}}(t) \subseteq V(t-1) \otimes V(t),$$

Obviously, the set of temporal edges of time slice $t$ is a subset of $E^{\text{int}}(t)$.

The subset $\text{int}(t) \subseteq V(t)$ is called the *interface of time slice $t$* and is defined as

$$\text{int}(t) = \{\alpha \in V(t) \mid \beta \in V(t-1), \{\alpha,\beta\} \in E^{\text{int}}(t)\}.$$

The moralized graph of the sample DPN DAG in Figure 1 appears in Figure 3, where the interfaces are indicated by filled circles (note that $\text{int}(0) = \emptyset$).

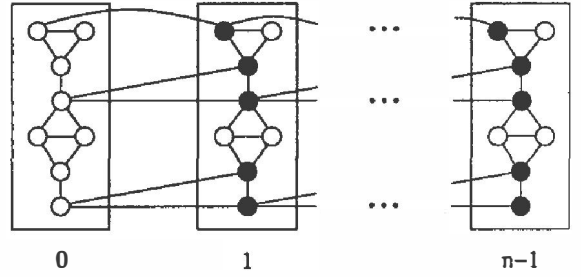

Figure 3: Sample initial DPN moral graph.

At any point in time, there is a series $\mathcal{P}_1,\ldots,\mathcal{P}_N$ of distinct but strongly related models, where each $\mathcal{P}_n$, $1 \leq n \leq N$, is specified by the quadruple $(p, \mathcal{G}_n, t_\pi(n), t_\phi(n))$, where $t_\pi(n) < t_\phi(n)$ is the oldest and $t_\phi(n)$ the newest time slice represented by $\mathcal{P}_n$, and where $\mathcal{G}_n = (V_n, E_n)$ is the independence graph of the probability $p$. At any time, $\mathcal{P}_N$ refers to the most recent model called the *current model*.

By the series $\mathcal{P}_1,\ldots,\mathcal{P}_N$ we understand the following. For any $1 \leq n \leq N$ the graph $\mathcal{G}_n$ of $\mathcal{P}_n$ is given by

$$(V_n, E_n) = \begin{cases} (V, E) & \text{if } n = N \\ (V \cup \text{int}(t'), \\ E \cup E^{\text{int}}(t')) & \text{if } n < N \end{cases} \quad (3)$$

where $t' = t_\pi(n+1)$, and $V$ and $E$ are given by (1) and (2), respectively, with $t - \pi = t_\pi(n) < t + \phi = t_\phi(n)$. Although $\mathcal{P}_n$, $n < N$, contains variables of $\mathcal{P}_{n+1}$ we define $t_\phi(n) = t_\pi(n+1) - 1$. Thus $t_\phi(n)$ represents the latest time slice about which $\mathcal{P}_n$ is guaranteed to be capable of containing complete information. For any $1 \leq n < N$, $t_\pi(n)$ and $t_\phi(n)$ are fixed. Also $t_\pi(N) = t_\phi(N-1) + 1$ is fixed, but $t_\phi(N)$ is a non-decreasing number meaning that the expanded model generated by including new time slices to $\mathcal{P}_N$ is still referred to as $\mathcal{P}_N$.

Finally, by

$$\mathcal{G}^N = \bigcup_{n=1}^{N} \mathcal{G}_n = \left( \bigcup_{n=1}^{N} V_n, \bigcup_{n=1}^{N} E_n \right)$$

we denote the *composite graph* of $\mathcal{G}_1,\ldots,\mathcal{G}_N$.

## 3 REASONING IN DPNs

The time slices $t_\pi(N),\ldots,t_\phi(N)$ of the current model, $\mathcal{P}_N$, are divided into two groups: the first $w$ slices constitute a group referred to as the *window of time slices* (or simply the *window*) and the remaining time slices comprising $t_\pi(N)+w, \ldots, t_\phi(N)$ are referred to as the *forecasting slices*; see Figure 4. Similarly, the time slices $0, \ldots, t_\phi(N-1)$ are referred to as the *backward smoothing slices*. Note that the term *forecasting slices* is slightly imprecise since all inference concerning variables of time slices for which no observations have been entered, actually are forecasts, even if such time slices belong to the window. For similar reasons the term *backward smoothing slices* is also slightly imprecise.

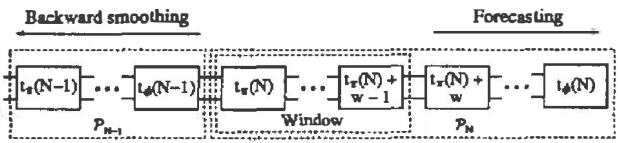

Figure 4: The current model, $\mathcal{P}_N$, includes a window of $w$ time slices.

For the purpose of making inferences, the window is assumed to consist of a triangulated version of the composite graph of the time slices involved. Hence a junction tree is associated with the window such that inferences in it are carried out as in a conventional static network. Inferences involving backward smoothing and forecasting are described in Sections 3.4 and 3.5.

The process of moving the window forward involves the two more or less separate processes of *model expansion* and *model reduction* discussed in detail in Sections 3.2 and 3.3. Since the window is represented by a junction tree, these processes roughly amount to, respectively, adding a new subtree to the junction tree and cutting off a part of the tree.

Model expansion by, say, $k$ new time slices consists of (a) adding $k$ new slices (conditional representation) to the current model (i.e. $t_\phi(N) := t_\phi(N) + k$), (b) moralizing the hybrid composite graph of the triangulated graph of the window and the DAGs of the $k$ consecutive time slices starting at $t_\pi(N) + w$, (c) triangulating that graph and identifying the new clique set, (d) constructing the new (expanded) junction tree, and (e) calibrating the new clique potentials with appropriate consideration of the old ones. As discussed in Section 3.2, the last step is optional. Expanding the current model by $k$ new time slices causes the width of the window to be increased by $k$, while the number of forecasting slices remains unchanged.

Model reduction by $k$ time slices involves elimination of all variables pertaining to time slices $t_\pi(N), \ldots, t_\pi(N) + k - 1$. Recall that elimination of a vertex $\alpha$ (variable $X_\alpha$) forms a complete subset of the vertices $\text{adj}(\alpha)$ unless they already constitute a complete set. The end-product of the elimination process is a potential involving the variables $\text{int}(t_\pi(N) + k)$. This potential, represented in one of the cliques of the reduced junction tree, $\Upsilon_N$ say, represents all information about the past necessary for the reduced model to take full account of the knowledge about the history of the system. Reducing the current model by $k$ time slices causes the number of backward smoothing slices to be increased by $k$ and the width of the window to be decreased by $k$, while the number of forecasting slices remains unchanged.

Two issues are of major importance here: (a) if backward smoothing is to be performed, the cliques of the triangulated graph resulting from the reduction process must be linked together in a new junction tree, $\Upsilon_{N-1}$ say, such that backward smoothing can be performed by passing messages from $\Upsilon_N$ to $\Upsilon_{N-1}$ via the potential involving variables $\text{int}(t_\pi(N) + k)$, and (b) since both the expansion and the reduction process performs a triangulation (i.e. finds an elimination order) of (basically) the same model, these two processes should be coordinated such that the same elimination order is employed.

The triangulation carried out as a subtask of the expansion process is unconstrained in the sense that the search space of elimination orders consists of all permutations of the set $V$ of vertices of the (expanded) window, whereas the reduction process may be perceived as a constrained triangulation, where the vertices eliminated define the prefix of orders comprising all vertices in $V$. Then obviously it might be advantageous to make a constrained decomposition in the first place, rendering the reduction process trivial, provided it is carried out in the fundamental way described above (i.e. assuming the reduction concerns $k$ lumps of $\mathcal{P}_N$, where each lump includes all vertices of a particular time slice). This introduces the notion of a *constrained elimination order* which is discussed further in Section 3.1.

### 3.1 CONSTRAINED ELIMINATION ORDERS

A constrained elimination order is defined as follows.

**Definition 1** *Let $\mathcal{G}^N = \bigcup_{n=1}^{N} \mathcal{G}_n = (V, E)$ be a composite graph and let $\# : V \mapsto \{1, \ldots, |V|\}$ define an elimination order. This order is said to be constrained if $\#(\alpha) < \#(\beta)$ for all $1 \leq i < j \leq N$, $\alpha \in V_i$ and $\beta \in V_j$. Similarly, $T(\mathcal{G}_\#^N)$ is said to be a constrained triangulation of $\mathcal{G}^N$.*

Constrained elimination orders have a number of important properties which shall be used in Sections 3.3 and 3.4.

First, we observe that the order in which the vertices $\bigcup_{0 < n < N} V_n$ are eliminated does not affect the complexity of $\mathcal{P}_N$. This fact follows from Lemma 1 the proof of which has been made by Rose, Tarjan and Lueker (1976).

**Lemma 1 (Rose et al. (1976))** *Let $\mathcal{G}_\# = (V, E)$ be an ordered graph. Then $\{\alpha, \beta\} \in E \cup T(\mathcal{G}_\#)$ if* and only if there is a path $\langle \alpha = \alpha_1, \ldots, \alpha_k = \beta \rangle$ such that $\#(\alpha_i) < \min\{\#(\alpha), \#(\beta)\}$ for all $1 < i < k$.

This property implies that, under constrained elimination, an optimal elimination order for $\mathcal{G}^N = \bigcup_{n=1}^{N} \mathcal{G}_n$ is given by optimal orders for $\mathcal{G}_n$, $1 \leq n \leq N$.

**Lemma 2** Let $\mathcal{P}_1, \ldots, \mathcal{P}_N$ be a series of conditional models with composite moral graph $(\mathcal{G}^N)^m$, and let $\mathcal{P}_1^*, \ldots, \mathcal{P}_N^*$ be the corresponding constrainedly decomposable models with composite graph $(\mathcal{G}^*)^N$. Then for any $1 \leq t \leq t_\phi(N)$, $\text{int}(t)$ in $(\mathcal{G}^N)^m$ is a complete separator of $(\mathcal{G}^*)^N$.

Proof: From the definition of $\text{int}(t)$ it follows that $\text{int}(t)$ is a separator of $(\mathcal{G}^N)^m$. Since $\mathcal{P}_1^*, \ldots, \mathcal{P}_N^*$ are constrainedly decomposable it follows from Lemma 1 that for all paths $\langle \alpha = \alpha_1, \ldots, \alpha_k = \beta \rangle$, where $\alpha \in V(t-1)$ and $\beta \in V(t) \setminus \text{int}(t)$, $\{\alpha_1, \ldots, \alpha_k\} \cap \text{int}(t) \neq \emptyset$. That is, $\text{int}(t)$ is also a separator of $(\mathcal{G}^*)^N$. Also due to the constrained elimination order it follows from Lemma 1 that $\text{int}(t)$ induces a complete subgraph of $(\mathcal{G}^*)^N$. □

Thus under constrained elimination the interface of time slice $t$, $1 \leq t \leq t_\phi(N)$, is identical in the moral and the corresponding decomposable graphs. This result is used in the following.

**Lemma 3** Let $\mathcal{P}_1, \ldots, \mathcal{P}_N$ be a series of constrainedly decomposable models with composite graph

$$\mathcal{G}^N = \bigcup_{n=1}^{N} \mathcal{G}_n = (V, E).$$

Then $\mathcal{G}^N$ is constrainedly triangulated.

Proof: From Lemma 2 we have that for any $1 \leq t \leq t_\phi(N)$, $\text{int}(t)$ is a complete separator of $\mathcal{G}^N$ and hence $(A, B, \text{int}(t))$ is a decomposition of $\mathcal{G}^N$, where $A = V(1) \cup \cdots \cup V(t-1)$ and $B = V \setminus (A \cup \text{int}(t))$. The graphs $\mathcal{G}^N_{A \cup \text{int}(t)}$ and $\mathcal{G}^N_{B \cup \text{int}(t)}$ have complete separators $\text{int}(1), \ldots, \text{int}(t)$ and $\text{int}(t), \ldots, \text{int}(t_\phi(N))$, respectively. Continuing this argument we end up with subgraphs $\mathcal{G}_1, \ldots, \mathcal{G}_N$ all of which are constrainedly triangulated, and the result follows. □

This shows that backward smoothing, at least in principle, can be accomplished by constructing a junction tree for $\mathcal{G}^N$ and performing propagation in that tree. However, a less space consuming technique exists as described in Section 3.4.

### 3.2 MODEL EXPANSION

The operation of expanding the current model by, say, $k$ new time slices $t_\phi(N) + 1, \ldots, t_\phi(N) + k$ is carried out for the purpose of including $k$ new time slices (not necessarily $t_\phi(N)+1, \ldots, t_\phi(N)+k$) into the window. The wish to expand the window may be explicit or implicit as part of the operation of moving the window $k$ time slices forward.

A new time slice is added to the current model via conditional probability relations such that the variables added have parents among the variables of the current model (relations in the opposite direction are not allowed). The structure of the DAGs of the conditional models of individual time slices will most often be identical. Note, however, that we make no structural or logical restrictions as to the conditional networks and temporal relations added. Thus, if an initial assumption implying identical time slice models turns out to be inadequate or erroneous, the presented scheme poses no obstacles to changing such assumptions.

In order to produce a junction tree for the expanded window we perform the operations of moralization and triangulation. The moralization step involves moralization of the hybrid composite graph (of the triangulated graph of the window and the DAGs of the $k$ new time slices) and implies that the conditional probabilities of the $k$ new time slices of the window are conceived as potentials. These potentials are in turn attached to appropriate cliques of the triangulated graph resulting by employing the constrained triangulation scheme to the moralized graph. A sample model expansion is shown in Figure 5, where the dashed lines are the edges added by moralization. In this example, the window is assumed to consist of a single time slice (the initial one).

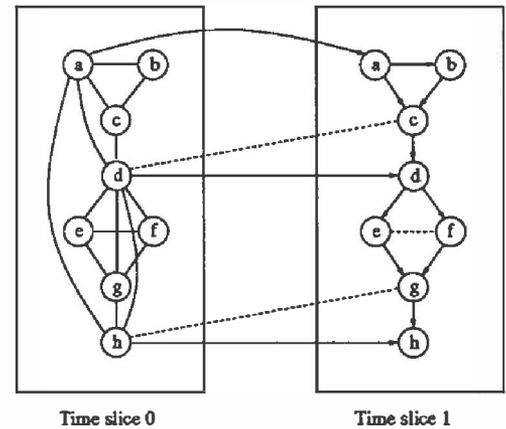

Time slice 0        Time slice 1

Figure 5: Sample model expansion.

Obviously, in finding an optimal elimination order, we have to take into consideration the topology of the graph as it appears after addition of the next time slice. Since we want the model complexity in terms of the state space size to be as low as possible to minimize the complexity of inference, and since the state space size varies heavily over the range of elimination orders, a careful analysis must be conducted to establish an appropriate order. To find an optimum elimination order for an arbitrary graph is, however, an NP-hard problem as proved by Wen (1990). Yet, in practice it turns out that near optimum triangulations may be found using simple heuristic ordering strategies (Rose 1973, Kjærulff 1992). In Figure 5 the applied elimination order is $b, e, f, c, g, d, a, h$. (The original directed and moral graphs are shown in Figures 1 and 3.)

Having found the cliques of the new expanded graph on the basis of an appropriate elimination order, the next step concerns construction of a junction tree for those cliques. As much as possible of the junction tree, $\Upsilon = (\mathcal{C}, \mathcal{E})$, in existence prior to the expansion should be reused in order to minimize the amount of work required to construct the expanded junction tree $\Upsilon' = (\mathcal{C}', \mathcal{E}')$. Note that as a direct consequence of the constrained decomposition scheme there is for each 'old' clique $C \in \mathcal{C}$ a 'new' clique $C' \in \mathcal{C}'$ such that $C \subseteq C'$. For some cliques the containment might be strict. The creation of $\Upsilon'$ can be described as follows.

1. Identify the set $\mathcal{C}'$ of cliques of $\Upsilon'$.

2. Construct a 'skeleton' of $\Upsilon'$:

    (a) Create clique objects for all members of $\mathcal{C}' \setminus \mathcal{C}$ and clique intersection objects for all members of $\mathcal{E}' \setminus \mathcal{E}$.

    (b) Initiate the potential tables of these new clique and clique intersection objects to unity. (The potential tables of the cliques in $\mathcal{C} \cap \mathcal{C}'$ and of the clique intersections in $\mathcal{E} \cap \mathcal{E}'$ remain unchanged.)

3. For each $C \in \mathcal{C} \setminus \mathcal{C}'$ and each $E \in \mathcal{E} \setminus \mathcal{E}'$ (i.e. 'old' cliques rendered redundant and their associated intersections) attach (by multiplication) the associated potential tables to the tables of appropriate clique and clique intersection objects.

4. Attach the conditional probability tables of the variables of the new time slices to appropriate new cliques.

(The term 'appropriate' in points 3 and 4 refers to the index set of the table to be attached being a subset of the clique or clique intersection upon which it is attached.) The expanded junction tree $\Upsilon'$ has now been created. That is, a potential representation for the joint probability distribution for the expanded window has been established. In Figure 6 the cliques and clique intersections remaining unchanged are shown in bold and the attachment of potential tables of redundant 'old' cliques and clique intersections are indicated by dashed arrows. Note that the cliques has been numbered according to the order of creation using the above elimination order and that clique 5 in part a is a proper subset of clique 5 in part b.

Now, if we have an immediate interest in the marginal distributions of variables (or sets of variables) in the $k$ new time slices of the window, a propagation can be performed; otherwise we might postpone the propagation step until e.g. new observations has been recorded. If $\Upsilon$ was calibrated immediately before the model expansion was executed, we only need to perform prop-agation in the subtree induced by the set of new cliques.

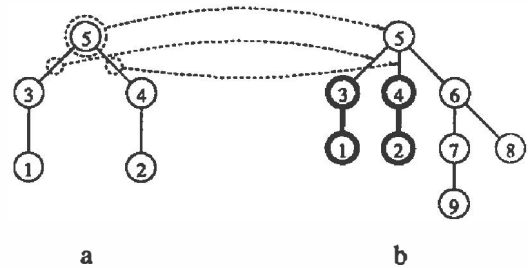

Figure 6: Sample junction tree expansion

### 3.3 MODEL REDUCTION

Due to the constrained decomposition scheme employed by the model expansion process, model reduction becomes a relatively easy task as previously discussed. In developing a model reduction scheme it is important to recognize the requirements for convenient backward smoothing beyond time slice $t_\pi(N)$. Below we develop a reduction scheme which meets such requirements and which is based on the results of the following theorem.

**Theorem 1** Let $\mathcal{P}_1, \ldots, \mathcal{P}_N$ be a series of constrainedly decomposable models, where each $\mathcal{P}_i$, $1 \leq i \leq N$, is calibrated. Assume $\mathcal{P}_{n-1}$ and $\mathcal{P}_n$ are jointly uncalibrated for some $1 < n \leq N$. Complete information required to calibrate $\mathcal{P}_{n-1}$ to $\mathcal{P}_n$ or vice versa is represented by the marginal $\psi_{int(t_\pi(n))}$, where there is a clique $C_1$ of $\mathcal{G}_{n-1}$ and a clique $C_2$ of $\mathcal{G}_n$ such that $int(t_\pi(n)) \subset C_1$ and $int(t_\pi(n)) \subseteq C_2$.

*Proof:* From Lemma 2 we have that $int(t_\pi(n))$ is a complete separator of $\mathcal{G}_{n-1} \cup \mathcal{G}_n$ and hence $\psi_{int(t_\pi(n))}$ contains complete mutual information between $\mathcal{P}_{n-1}$ and $\mathcal{P}_n$. From the definition of $\mathcal{G}_i$, $1 \leq i < N$, (cf. (3)) we have that $int(t_\pi(n)) \subset V(t_\phi(n-1))$, and since for each pair $\{\alpha, \beta\}$, where $\alpha \in V(t_\phi(n-1))$ and $\beta \in int(t_\pi(n))$, $\#(\alpha) < \#(\beta)$, $int(t_\pi(n))$ induces a complete subgraph of $\mathcal{G}_{n-1}$. Hence there is a clique $C_1$ of $\mathcal{G}_{n-1}$ such that $int(t_\pi(n)) \subset C_1$. Since $int(t_\pi(n))$ is complete in $\mathcal{G}_{n-1}$ it follows immediately that it is also complete in $\mathcal{G}_n$ and hence there is a clique $C_2$ in $\mathcal{G}_n$ such that $int(t_\pi(n)) \subseteq C_2$. □

So far we have not been concerned with the process of creating new models to be added to a series $\mathcal{P}_1, \ldots, \mathcal{P}_N$. However, the reduction process partition $\mathcal{P}_N$ into two models, one representing the time slices eliminated and the other the remaining time slices of $\mathcal{P}_N$ (subsequently defining the new current model). That is, whenever $\mathcal{P}_N$ is subjected to reduction, the number, $N$, of models is increased by one. Thus, conforming to (3), we define the reduction of $\mathcal{P}_N$ by the $k$ oldest time slices by sequentially executing the following steps.

1. Let $\mathcal{P}' = (\mathcal{G}' = (V', E'), t_0 = t_\pi(N), t_k = t_\pi(N) + k)$, where $0 \leq k < t_\phi(N) - t_\pi(N)$ and

$$V' = V(t_0) \cup \cdots \cup V(t_k) \cup \text{int}(t_k + 1),$$
$$E' = E(t_0) \cup E^*(t_1) \cup \cdots \cup E^*(t_k).$$

2. Let $N := N + 1$.

3. Let $\mathcal{P}_N = (\mathcal{G}_N = (V_N, E_N), t_\pi(N) = t_k + 1, t_\phi(N) = t_\phi(N-1))$, where

$$V_N = V_{N-1} \setminus (V' \setminus \text{int}(t_\pi(N))),$$
$$E_N = E_{N-1} \setminus E'.$$

4. Let $\mathcal{P}_{N-1} = \mathcal{P}'$.

In terms of operations on the junction tree of $\mathcal{P}_N$ (actually the junction tree of the window) an equivalent description of the reduction process may be formulated as follows, where $t = t_\pi(N) + k + 1$ is the oldest time slice of the window when the reduction has been completed.

1. Prior to the reduction, let $\Upsilon = (\mathcal{C}, \mathcal{E})$ be a junction tree for $\mathcal{P}_N$.

2. Let $\mathcal{C}' = \{C \in \mathcal{C} \mid C \cap \bigcup_{i=t_\pi(N)}^{t_\pi(N)+k} V(i) \neq \emptyset\}$ be the cliques containing variables to be eliminated, and $\mathcal{C}'' = \mathcal{C} \setminus \mathcal{C}'$ the remaining cliques.

3. Let $\Upsilon' = \Upsilon_{\mathcal{C}'}$ and $\Upsilon'' = \Upsilon_{\mathcal{C}''}$ be the junction trees induced by $\mathcal{C}'$ and $\mathcal{C}''$, respectively (see Figure 7).

4. Let $\mathcal{B} = \{C \in \mathcal{C}'' \mid \text{adj}(C) \cap \mathcal{C}' \neq \emptyset \text{ in } \Upsilon\}$.

5. If there is no $C \in \mathcal{B}$ such that $\text{int}(t) \subseteq C$ then add $\text{int}(t)$ to $\mathcal{C}''$ and let $\text{adj}(\text{int}(t)) = \mathcal{B}$; otherwise add $\mathcal{B} \setminus \{C\}$ to $\text{adj}(C)$.

6. Let $N := N + 1$.

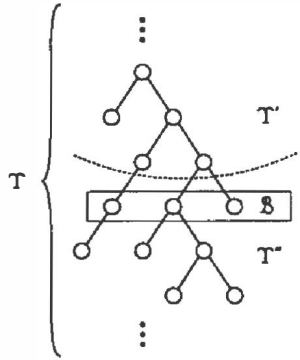

Figure 7: Partitioning $\Upsilon$ into $\Upsilon'$ and $\Upsilon''$.

After the execution of Steps 1–6, $\mathcal{P}_{N-1}$ is given by $\Upsilon'$ and $\mathcal{P}_N$ by $\Upsilon*$ which is the result of modifying $\Upsilon''$ as described in Step 5 above. It is easily verified that $\Upsilon*$ is a junction tree for $\mathcal{G}_N$ of Step 3 of the four-step description of the reduction process.

First assume that the condition of the 'if' part of Step 5 holds. Since the constrained decomposition forces $\text{int}(t)$ to induce a complete subgraph of $\mathcal{G}_N$ and since there is no clique in $\mathcal{C}''$ containing $\text{int}(t)$, then $\text{int}(t)$ itself must be a clique of $\mathcal{G}_N$. The subset $\mathcal{B} \subseteq \mathcal{C}''$, where for each $B \in \mathcal{B}$ there is a non-empty intersection between the adjacency set of $B$ and $\mathcal{C}'$ in $\Upsilon$, is then made the adjacency set of $\text{int}(t)$. Since the path in $\Upsilon$ between any pair of elements of $\mathcal{B}$ includes elements of $\mathcal{C}'$ (i.e. $\mathcal{C}'$ separates the elements of $\mathcal{B}$ from one another), this does not violate the tree structure of $\Upsilon*$. Neither does it violate the property of $\Upsilon*$ being a junction tree, as the intersection of any pair $(C', C'')$ of cliques, where $C' \in \mathcal{C}'$ and $C'' \in \mathcal{C}''$, is a subset of $\text{int}(t)$.

Next, assume the condition to fail (i.e. there is a clique $C \in \mathcal{C}''$ such that $\text{int}(t) \subseteq C$) in which case $\mathcal{B} \setminus \{C\}$ is made a subset of the adjacency set of $C$ in $\Upsilon''$. With arguments similar to those above it is readily realized that the property of $\Upsilon*$ being a junction tree is not violated.

### 3.4 BACKWARD SMOOTHING

Clearly, the arrival of external evidence (observations) affects not only the estimates of (unobserved) variables of the relevant time slice(s), but may also have significant effect on estimates of variables of other time slices. The process of re-estimating variables of past slices in light of new evidence (retrospective assessment) is often referred to as *backward smoothing*. If the variables for which re-estimated probability distributions are required, are all included in the current model, $\mathcal{P}_N$, backward smoothing is an implicit part of propagation in the window of time slices. However, if we want to backward smooth from $\mathcal{P}_N$ to $\mathcal{P}_{N-1}$ special actions should be taken. Specifically, complete information about observations pertaining to the window should be transferred from $\mathcal{P}_N$ to $\mathcal{P}_{N-1}$.

Given the model reduction strategy described in Section 3.3 the process of propagating complete relevant information backward from $\mathcal{P}_n$ to $\mathcal{P}_{n-1}$ or forward from $\mathcal{P}_{n-1}$ to $\mathcal{P}_n$ becomes very simple. Consider the example where $\mathcal{P}_1, \ldots, \mathcal{P}_N$ are calibrated, but jointly uncalibrated. Let the inconsistency be caused by a series $E_2, \ldots, E_N$ of sets of external evidence, such that $\mathcal{P}_n$, $1 \leq n \leq N$, is uninformed of $E_{n+1}, \ldots, E_N$. Now, $\mathcal{P}_n$ may become informed of $E_{n+1}, \ldots, E_N$ by the following calibration process (see also Figure 8). For convenience we first define the concept of an *interface clique* as follows.

**Definition 2** Let $\mathcal{P}_1, \ldots, \mathcal{P}_N$ be a series of constrainedly decomposable models. Then for any $1 \leq n \leq N$ let $\mathcal{I}_n^-$ denote the set of cliques of $\mathcal{G}_n$ such that for any $IC_n^- \in \mathcal{I}_n^-$, $\text{int}(t_\pi(n)) \subseteq IC_n^-$. Similarly, for any $1 \leq n < N$ let $\mathcal{I}_n^+$ denote the set of cliques of $\mathcal{G}_n$ such that for any $IC_n^+ \in \mathcal{I}_n^+$, $\text{int}(t_\pi(n+1)) \subset IC_n^+$. $IC_n^-$ and $IC_n^+$ are called interface cliques of $\mathcal{P}_n$.

1. Initially let $i = N$. Then repeat steps 2 and 3 sequentially while $i > n$.

2. Let $IC_i^- \in \mathcal{I}_i^-$, $IC_{i-1}^+ \in \mathcal{I}_{i-1}^+$, and $I = \text{int}(t_\pi(i))$.

$$\psi_{IC_{i-1}^+}^* = \psi_{IC_{i-1}^+} \frac{\sum_{IC_i^- \setminus I} \psi_{IC_i^-}}{\sum_{IC_{i-1}^+ \setminus I} \psi_{IC_{i-1}^+}},$$

where superscript "*" denotes the updated potential.

3. Calibrate $\mathcal{P}_{i-1}$ by propagation and decrement $i$ by one.

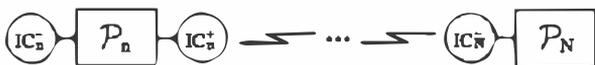

Figure 8: Backward smoothing from $\mathcal{P}_N$ to $\mathcal{P}_n$.

## 3.5 FORECASTING

In time-series analysis applications there is typically a desire to make optimal forecasts of the random process considered. Within the computational framework presented above, forecasts which do not exceed the extent of the window are an implicit part of propagation in the junction tree of the window; otherwise it may be performed by expanding the window by the required number time slices. If forecasts are wanted for a large number of time slices ahead of the window, the complexity of the resulting decomposable model might, however, easily exceed the capacity of the available computing resources. Such cases may be solved in a number of ways.

One is to move the window the required number of steps, where propagation is performed in each step, and subsequently, moving it back again. This might, however, be a very time consuming operation, and furthermore, a lot of unnecessary calculations will quite often be carried out as we typically only want the forecasts for a limited number of variables. Therefore, there is a demand for alternative forecasting methods which either avoids the junction tree approach and/or exploits the fact that forecasts are only required for a limited number of variables.

Concerning non-junction-tree methods (i.e. no triangulation), various Monte-Carlo sampling schemes may be useful. A common trait of these schemes is the fact that the variance of the resulting distributions can be made arbitrarily small. In fact, some of the most fruitful approaches to variance reduction is Monte-Carlo sampling (Ripley 1987). Note, however, that a reduction of the standard error of an estimator by a factor of $k$ requires an increase in the sampling size, $n$, by around a factor of $k^2$ due to the ubiquitous $1/\sqrt{n}$ law of statistical variation. Thus, to get forecasts within a small distance from the 'exact' values, we should expect the computing time to be relatively large; in some cases even larger than those required by exact methods, but of course with much less space requirements since the sampling is performed in the DAG structure involving relatively low-dimensional probability tables. Another important feature of sampling methods is that the time complexity grows only linearly in the dimensionality of the tables involved, whereas it grows exponentially for exact methods.

Another method that might be fruitful is based on the fact that (a subset of) the conditional probabilities of a probabilistic model quite often exhibits linearity in the sense that they are (approximately) linear functions in the variables upon which they are given. That is,

$$p(x_\alpha) \approx \sum_{\mathcal{X}_{\text{pa}(\alpha)}} p(x_\alpha \mid x_{\text{pa}(\alpha)}) \prod_{\beta \in \text{pa}(\alpha)} p(x_\beta).$$

The method is then simply given by calculating all such approximate marginal probability distributions in an appropriate order (i.e. the distributions of all parents of a variable should be calculated before the distribution of the variable itself). Given that the divergence between such approximate distributions and the 'exact' ones are below an acceptable upper bound for the variables of interest, this is a very fast forecasting method. The interesting point concerning the exactness of the method is that an upper bound on the error can be computed in advance by application of theorems of linear algebra.

## 4 SUMMARY

We have presented a computational scheme for reasoning in dynamic probabilistic networks featuring description of non-linear, multivariate dynamic systems with complex conditional independence structures and providing a mechanism for efficient backward smoothing. As opposed to a static network representing a finite and fixed number of time slices (i.e. capable of reasoning only about a finite series of observations of a dynamic system) the proposed scheme can handle infinite series of observations. Further, in applying static networks representing a fixed number of time slices as models of dynamic systems, there is typically a desire to include as many time slices as possible in the model. Thus, inference easily becomes time consuming and inflexible (i.e. propagation involves all time slices in the model even if updated distributions are wanted only for a limited number of time slices). The proposed scheme, on the other hand, provides a high degree of flexibility in the reasoning process, since the width of the window of time slices can be changed dynamically as well as the number of 'backward smoothing slices' and the number of 'forecasting slices'. In addition, the scheme provides selective inference in the sense that inference can be performed in (i) the window, as (ii) backward smoothing, or as (iii) forecasting.

Since the presented model reduction scheme supports a convenient and efficient backward smoothing method it also supports inclusion and modification of observations pertaining to time slices 'to the left of' the window. Delayed observations is a quite typical phenomenon; for example, in a medical setting delays may be caused by processing time in a laboratory (e.g. analysis of a blood sample).

Although we have presented a scheme for reasoning in dynamic networks, a range of issues still remain to be dealt with. A couple of the most important issues are the following.

Only preliminary studies has been carried out to investigate the applicabilities the various forecasting methods discussed in Section 3.5. Especially, a scheme for establishing an upper bound on the forecast error by applying the linear approximation algorithm is desirable. But also a study of the applicability of various Monte-Carlo sampling schemes should be conducted.

Since many applications feature a large number of temporal relations, the state space sizes of the interface cliques of the time slices of the window and of the 'backward smoothing slices' may become unmanageably large. In such cases there will be a need for approximations. One obvious way of approximating the inference is to exclude some of the edges required between members of the interface set of a time slice. An extreme approach could be assumption of independence between all parents of interface variables (i.e. no fill edges at all added between interface vertices). To that end, studies on the upper bounds of the resulting error and its attenuation as time evolves, should be conducted.

An implementation of the computational scheme presented in this paper has been built on top of the HUGIN shell.

### Acknowledgements

I wish to thank Steffen L. Lauritzen for his valuable comments on an earlier draft of this paper and other members of the ODIN group at Aalborg University for stimulating discussions.